\documentclass[hyphens]{article} 
\usepackage{iclr2027_conference,times}

\usepackage{amsmath,amsfonts,bm}

\def\eqref#1{equation~\ref{#1}}

\def\1{\bm{1}}

\DeclareMathAlphabet{\mathsfit}{\encodingdefault}{\sfdefault}{m}{sl}
\SetMathAlphabet{\mathsfit}{bold}{\encodingdefault}{\sfdefault}{bx}{n}

\usepackage{hyperref}
 \usepackage{wrapfig}
\usepackage{url}  
\usepackage{graphicx} 
\usepackage{natbib}  
\usepackage{caption} 
\usepackage{makecell}
\usepackage{algorithm}
\usepackage{algorithmic}
\usepackage{colortbl}       
\usepackage{newfloat}
\usepackage{listings}
\DeclareCaptionStyle{ruled}{labelfont=normalfont,labelsep=colon,strut=off} 
\floatstyle{ruled}
\newfloat{listing}{tb}{lst}{}
\floatname{listing}{Listing}

\usepackage{booktabs}

\usepackage{multirow} 
\usepackage{amsmath}
\usepackage{amssymb}
\usepackage{amsfonts} 
\usepackage[utf8]{inputenc}
\usepackage[T1]{fontenc}
\usepackage{nicefrac}
\usepackage{microtype} 
\usepackage{colortbl}
\usepackage{array}
\usepackage{pifont} 
\usepackage{xspace}
\usepackage{cleveref}
\usepackage{enumerate}
\usepackage{enumitem}
\usepackage{subcaption}
\usepackage{booktabs}
\usepackage{graphicx}
\newcommand{\method}{{\fontfamily{lmtt}\selectfont \textbf{Trace}}\xspace}
\usepackage[dvipsnames]{xcolor}
\usepackage[most]{tcolorbox}

\newcommand{\hlfirst}[1]{\colorbox[HTML]{CFE2FF}{#1}}  
\newcommand{\hlsecond}[1]{\colorbox[HTML]{E8F1FF}{#1}} 

\newcommand{\modelname}[1]{{\fontfamily{pcr}\selectfont {#1}}\xspace}
\newcommand{\methodname}[1]{{\fontfamily{pcr}\selectfont {#1}}\xspace}

\newcommand{\done}[1]

\newcommand{\blue}[1]{$_{\color{BlueGreen}\downarrow #1}$}
\newcommand{\red}[1]{$_{\color{RedOrange}\uparrow #1}$}
\definecolor{darksalmon}{rgb}{0.91, 0.59, 0.48}

\newcommand{\staticdelta}{\red{0.00}}

\title{Success Leaves Detours: Learning Executable Walkthroughs for Long-Horizon Agents}

\author{
Kaijie Chen\textsuperscript{1,\textdagger},
Chenyu Fang\textsuperscript{2,*},
Liang Yan\textsuperscript{3,\textdagger},
Bo Li\textsuperscript{3},
Bo Zhang\textsuperscript{2},
Peng Ye\textsuperscript{2,*}
\\[4pt]
\textsuperscript{1}Tongji University
\qquad
\textsuperscript{2}Shanghai AI Laboratory
\qquad
\textsuperscript{3}Fudan University
\\[3pt]
\textsuperscript{\textdagger}Equal contribution
\qquad
\textsuperscript{*}Corresponding authors
}

\iclrfinalcopy 
\begin{document}

\maketitle

\begin{abstract}
Test-time self-evolving agents improve by reusing experience across episodes, but sparse-reward trajectories contain failed attempts, loops, and detours that hinder direct reuse, while trajectory summaries often discard the state conditions and action dependencies required for execution. We study \emph{executable Walkthrough induction from sparse-reward trajectories}, aiming to extract compact, state-conditioned, and verifiable procedures. Our key observation is that delayed credit measures an action's association with downstream progress but does not establish whether the action produces a state fact required for subsequent execution. We therefore propose \method, a credit-guided and dependency-grounded framework that compiles noisy interaction histories into executable Walkthrough Memory. It first detects progress anchors from reward and persistent state changes and propagates credit backward to identify high-value transitions. It then estimates the prerequisites of each action from cross-episode success and failure evidence. Finally, backward dependency slicing traces anchor facts to their observed producers and extracts compact, dependency-consistent action chains, while filtering loops and detours that are not linked to the resolved dependencies. The resulting Walkthroughs explicitly encode entry conditions, ordered state--action--effect steps, completion and failure predicates, and prerequisite dependencies, enabling cross-episode reuse, intermediate-state resumption, and programmatic verification. Experiments on J-TTL, WebShop, and ScienceWorld with three open-source LLM backbones show that \method consistently outperforms eight test-time learning and memory baselines. Compared with the strongest baseline, it improves average AUC and Final-$3$ by $30.0\%$ and $40.5\%$, respectively, while using fewer inference tokens. Ablations and controlled studies validate the contributions of credit propagation, prerequisite estimation, dependency slicing, executable representation, and dependency-aware selection. These results suggest that long-horizon interaction is better supported by state-conditioned executable procedures than by complete trajectories or abstract summaries.
\end{abstract}

\section{Introduction}
\label{sec:intro}



Large Language Model (LLM) agents have demonstrated substantial progress on long-horizon interactive tasks, including web navigation, software operation, text-based games, and tool-augmented problem solving~\citep{zhou2024webarena,yao2022webshop,wang2022scienceworld}. Despite this progress, long-horizon environments remain difficult because they are often partially observable, governed by sparse or delayed feedback, and structured around prerequisite chains whose consequences emerge only many steps after an action is taken. Test-time self-evolving agents seek to overcome these limitations by accumulating interaction experience across episodes and updating external memories, prompts, or agent configurations without repeatedly modifying the underlying model~\citep{shinn2023reflexion,zhao2024expel,he2026evotest}. The resulting improvement, however, depends critically on whether noisy interaction trajectories can be transformed into memories that later Actors can reliably select, execute, and verify.

Consider a broad class of long-horizon environments in which meaningful progress is distributed across many steps, explicit reward is sparse or delayed, and successful execution depends on latent prerequisites. A trajectory that ultimately opens a gate may contain \texttt{GO SOUTH}, \texttt{GO NORTH}, \texttt{ASK TECHNICIAN}, \texttt{SEARCH DESK}, \texttt{TAKE CODE}, and \texttt{OPEN GATE}. Only the final action may receive reward, even though asking the technician, searching the desk, and taking the code establish indispensable conditions, while the back-and-forth navigation may be incidental. Memory construction in this setting must therefore recover unrewarded preparation, distinguish necessary steps from coincidental detours, identify coherent procedure boundaries, and preserve the state conditions under which each step remains valid. We formulate this problem as \emph{\textbf{executable Walkthrough induction from sparse-reward trajectories}}: recovering minimal, state-conditioned, executable, and verifiable procedures that can reproduce progress in subsequent episodes.

Existing agent memory systems approach experience reuse through four broad representational paradigms, each preserving a different aspect of interaction history while omitting part of the execution structure required in this setting. \textbf{(1) Episodic replay and retrieval methods} preserve or recall concrete past experience~\citep{huang2025r2d2,wang2026samem}; although such memories improve behavioral fidelity and state alignment, retrieval alone does not identify a minimally sufficient procedure, determine which recorded steps remain necessary, or specify where execution should resume and terminate. \textbf{(2) Semantic distillation and reflective memory} compress trajectories into reflections, insights, rules, or evolving playbooks~\citep{shinn2023reflexion,zhao2024expel,zhang2025ace}; these representations are compact and transferable, but they often abstract away explicit preconditions, intermediate state effects, resource-acquisition paths, and programmatic completion or failure criteria. \textbf{(3) Procedural and workflow-oriented memories} extract reusable routines or structured abstractions~\citep{cao2026reme,liang2026mcma,sun2026magnet}; however, their induction objectives do not explicitly test whether every retained transition establishes a fact required by a later action, so successful trajectories may still yield procedures that contain incidental steps or fail under altered intermediate states. \textbf{(4) Graph-structured memories} organize experience through temporal, semantic, hierarchical, or quasi-symbolic relations~\citep{zhang2026g,ranaldi2026evolving}; such connectivity supports structured retrieval and abstraction, but an edge is not an executable dependency unless the state produced upstream satisfies a precondition downstream. The common missing element is therefore a fact-grounded criterion that separates historical association with success from procedural necessity.

The central technical problem is \emph{\textbf{grounded procedure induction}}: identifying the smallest action chain that establishes a target state together with all prerequisites required to reach it. We address this problem with \method, a credit-guided and dependency-grounded framework for inducing executable Walkthroughs from noisy, sparse-reward trajectories. Its key insight is that delayed credit and procedural necessity provide complementary but non-equivalent evidence. Delayed credit can recover early actions that may contribute to later progress, including preparation with no immediate reward, but high credit only indicates historical relevance and does not prove that an action is required. Procedural necessity instead depends on whether an action establishes the target state or a prerequisite fact consumed by a later step. Accordingly, \method uses progress-aware credit propagation to obtain high-recall candidate actions, then applies fact-grounded prerequisite analysis to retain only the transitions needed to reproduce the target progress. Credit identifies potentially useful evidence, while dependency recovers the minimally sufficient execution chain.



\method converts raw interaction trajectories into executable Walkthroughs through a three-stage programmatic pipeline. First, it canonicalizes observations into structured states, detects progress anchors from rewards and persistent state changes, and propagates credit backward to recover potentially useful preparatory transitions. Second, it estimates prerequisite facts from observed action outcomes and cross-episode evidence, then performs required-fact-driven backward slicing to retain only transitions that establish the target state or an unresolved prerequisite. This removes loops, failed branches, redundant detours, and high-credit actions that are not procedurally necessary. Third, the retained dependency chain is compiled into a structured Walkthrough with explicit entry conditions, ordered precondition-action-effect steps, prerequisite relations, and completion and failure predicates. The resulting memory supports applicability checking, intermediate-state resumption, effect verification, and reliable termination. Across episodes, only the external Walkthrough Memory evolves, while the Actor remains fixed.

Empirically, \method delivers stronger and more consistent test-time improvement than memory, retrieval, reflection, and automated prompt-optimization baselines. Across WebShop, J-TTL, and ScienceWorld benchmarks with \modelname{Qwen3-32B}, \modelname{Mistral-Small-3.2-24B}, and \modelname{GPT-OSS-20B}, \method achieves the strongest AUC and Final-$3$ performance in every evaluated benchmark-backbone block. Relative to the strongest competing baseline in each setting, it improves average AUC by 30.0\% and Final-$3$ by 40.5\%, with average gains of 65.4\% in AUC and 135.6\% in Final-$3$ on the prerequisite-intensive J-TTL benchmark. Diagnostic ablations further establish a coherent mechanism-to-outcome chain: delayed credit increases preparatory-action recall, prerequisite estimation improves dependency recovery and necessary-step recall, and backward slicing raises necessary-step precision, reduces residual redundancy, and improves replay success. Robustness studies show gradual degradation under masked progress signals and injected exploration, loops, and backtracking, while increasing the history budget from three to nine episodes improves endpoint performance, despite non-monotonic intermediate results. These results support the central claim that effective long-horizon experience reuse requires both recovery of necessary state dependencies and an executable representation of the resulting procedure.

Our main contributions are as follows:

\begin{itemize}
    \item \textbf{Credit-Guided, Dependency-Grounded Induction.} We introduce \method for test-time self-evolution in long-horizon environments with sparse feedback, delayed consequences, hidden prerequisites, and noisy trajectories. By combining progress anchoring and delayed credit propagation with experience-grounded prerequisite estimation and required-fact-driven backward slicing, \method recovers unrewarded preparation while removing actions that are historically correlated with success but procedurally unnecessary.

    \item \textbf{Executable Self-Evolving Walkthrough Memory.} We propose a structured procedural memory that represents experience through entry conditions, prerequisite dependencies, ordered precondition-action-effect steps, and programmatic completion and failure predicates. This representation supports applicability checking, intermediate-state resumption, effect verification, reliable termination, and continual cross-episode refinement without updating the Actor parameters.

    \item \textbf{Cross-Benchmark Gains and Mechanistic Validation.} \method attains the strongest AUC and Final-$3$ results in every evaluated benchmark-backbone block, improving the strongest competing baselines by 30.0\% and 40.5\% on average and achieving particularly large gains on J-TTL. Fine-grained ablations, trajectory-quality diagnostics, progress-signal masking, noise injection, and history-efficiency studies consistently validate the distinct roles of credit propagation, prerequisite estimation, and dependency slicing.
\end{itemize}




\section{Related Work}
\label{sec:related}

\paragraph{Test-Time Learning and Self-Improving LLM Agents.} Existing work improves agents through reflection, online training, or system-level search. SAMULE~\citep{ge2025samule} generates reflections from individual interactions, within-task experience, and cross-task experience. WebEvolver~\citep{fang2025webevolver}, WebRL~\citep{qi2025webrl}, UI-Genie~\citep{xiao2026ui}, and Self-Challenging Agents~\citep{zhou2026self} convert interaction trajectories, generated data, or self-constructed tasks into training signals. Gödel Agent~\citep{yin2025godel} and AFlow~\citep{zhang2025aflow} optimize agents through self-modification or workflow search. These methods typically require parameter updates, reflection, or costly search. In contrast, we freeze both the model and agent framework and update only a cross-episode Walkthrough Memory through programmatic trajectory processing, enabling lightweight test-time improvement.

\paragraph{Procedural Memory and Workflow Induction from Agent Trajectories.} Evolving Generalist Virtual Agents~\citep{zhang2026evolving}, G-Memory~\citep{zhang2026g}, and Evolving Agents~\citep{ranaldi2026evolving} organize interaction experience into subtask graphs, hierarchical memories, or symbolic abstractions. ReMe~\citep{cao2026remember}, Learning How to Remember~\citep{liang2026learning}, and MAGNET~\citep{sun2026magnet} study the maintenance and transfer of procedural knowledge, while SAMem~\citep{wang2026samem}, R2D2~\citep{huang2025r2d2}, Stepwise Experience Recall~\citep{cui2025self}, and ExpSeek~\citep{zhang2026expseek} improve state- or step-level retrieval. These methods mainly reuse summaries, rules, or local trajectory fragments without explicitly representing step-level dependencies and execution conditions. Our method instead analyzes state transitions directly: progress anchors identify local goals, delayed credit recovers potentially useful preceding actions, and backward dependency slicing extracts compact, dependency-consistent action chains by retaining transitions linked to target facts and their candidate prerequisites. The resulting Walkthroughs contain entry conditions, ordered actions, expected effects, and completion and failure criteria, while a prerequisite DAG represents dependencies among them.

\paragraph{Exploration and Prerequisite-Aware Walkthrough Scheduling.} Existing exploration methods mainly optimize action, policy, or tool selection. Toward Efficient Exploration~\citep{arumugam2026toward} and Meta-RL~\citep{jiang2026metarl} study posterior sampling and learned exploration strategies. Progressive Exploration~\citep{qin2026learn}, Exploratory Memory-Augmented Agent~\citep{liu2026exploratory}, Dual-Scale World Memory~\citep{kim2026dual}, DreamPhase~\citep{hamidi2026dreamphase}, and SPORT~\citep{li2026iterative} improve exploration through self-imitation, memory, world models, offline imagination, or preference learning. MaestroMotif~\citep{klissarov2025maestromotif}, Agent S~\citep{agashe2025agent}, Cradle~\citep{tan2025cradle}, ToolTree~\citep{yang2026tooltree}, and Conductor~\citep{nielsen2026learning} search over predefined skills, tools, or agent sets, while COLA~\citep{zhao2025cola} and Cook and Clean Together~\citep{liang2026cook} study subtask scheduling. These approaches typically explore primitive actions or predefined skills and therefore cannot directly exploit prerequisite structures induced from prior trajectories. We schedule only Walkthroughs whose execution conditions are currently satisfied and propagate downstream returns through the prerequisite DAG, reducing ineffective exploration while capturing the long-term value of prerequisite procedures.




\begin{figure*}[t]
    \centering
    \includegraphics[width=\textwidth]{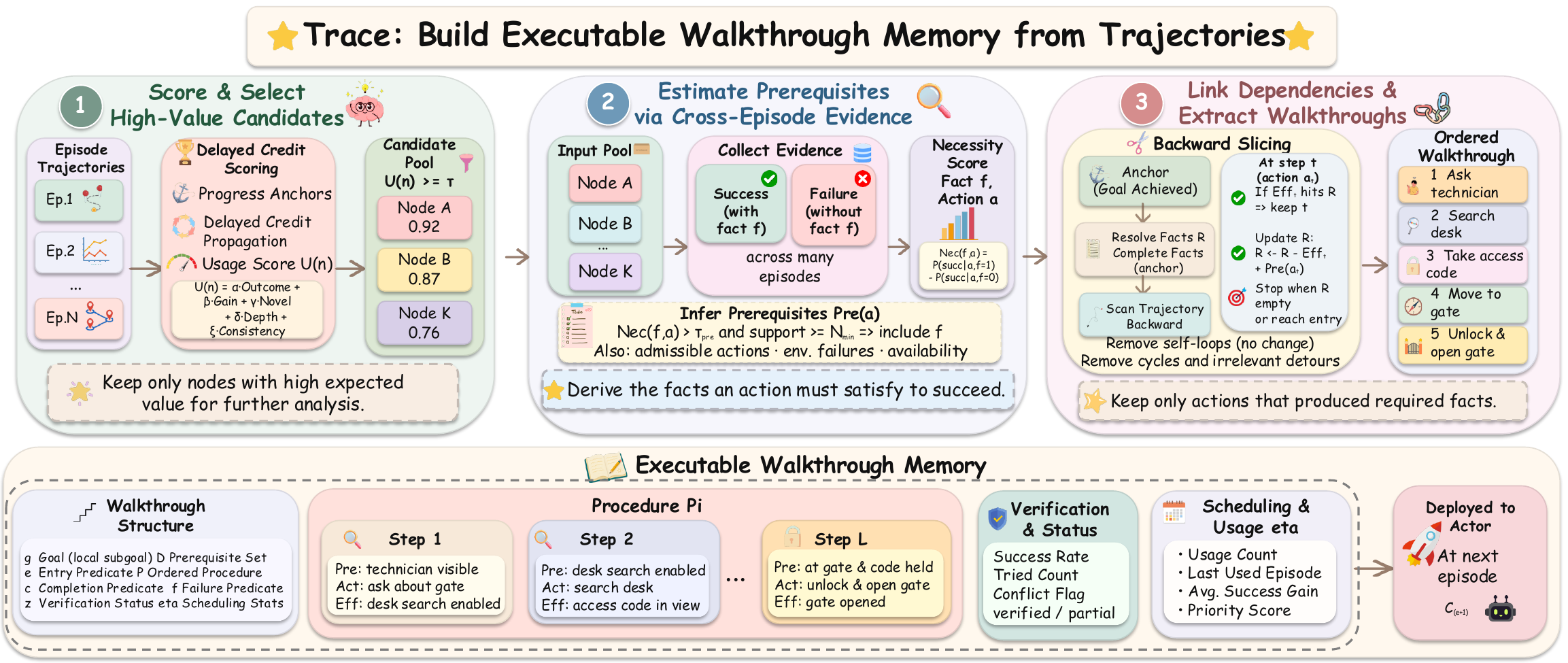}
    \caption{\textbf{Overview of \method.} The upper pipeline scores progress-linked candidates, estimates action prerequisites from cross-episode evidence, and slices trajectories backward from anchor facts. The lower panel shows the resulting executable Walkthrough, including state-conditioned steps, verification status, and scheduling statistics used by the next episode.}
    \label{fig:overview}
\end{figure*}

\section{Credit-Guided Walkthrough Memory Construction and Scheduling}
\label{sec:method}

 We consider a long-horizon interactive setting in which a frozen LLM Actor repeatedly attempts the same task across episodes. After each episode, the Evolver deterministically transforms the interaction trajectory into executable \emph{Walkthrough Memory}, where each Walkthrough encodes a reusable subgoal with its prerequisites, ordered steps, expected effects, and termination conditions. The Evolver detects progress anchors through state differencing, propagates delayed credit to earlier contributing actions, infers prerequisites from action effects and cross-episode evidence, and applies backward dependency slicing to extract compact,
dependency-consistent action chains from the observed trajectories.

\subsection{Progress Anchoring and Delayed Credit Propagation}
\label{sec:walkthrough_construction}

\paragraph{Canonical state representation.}
After an episode of $T$ actions, the Evolver receives
\begin{equation}
\tau=
{(o_t,a_t,r_t,\mathrm{inv}t)}{t=1}^{T},
\label{eq:trajectory}
\end{equation}
where $o_t$, $a_t$, $r_t$, and $\mathrm{inv}_t$ denote the observation, action, cumulative reward, and inventory at step $t$, respectively. Each observation is deterministically parsed into
\begin{equation}
s_t=
(\ell_t,\mathcal{I}_t,\mathcal{O}_t,
\mathcal{Z}_t,\mathcal{E}_t,\mathcal{A}_t),
\label{eq:canonical_state}
\end{equation}
where $\ell_t$ is the current location, $\mathcal{I}_t$ is the inventory, $\mathcal{O}_t$ is the visible-object set, $\mathcal{Z}_t$ contains object states, $\mathcal{E}_t$ contains exits, and $\mathcal{A}_t$ contains admissible actions.

The canonical state is assigned a stable key
$\kappa_t=\operatorname{Hash}(s_t)$. We define the reward and state changes as
$\Delta r_t=r_{t+1}-r_t$ and
$\Delta s_t=\operatorname{Diff}(s_t,s_{t+1})$, respectively. The trajectory then induces the following state--action transition:
\begin{equation}
e_t:\quad
\kappa_t
\xrightarrow{,a_t,}
\kappa_{t+1}.
\label{eq:compact_transition}
\end{equation}
Each transition is associated with the reward change $\Delta r_t$ and state change $\Delta s_t$. This representation exposes the persistent state changes and repeated transitions required by the subsequent credit and dependency analyses.

\paragraph{Progress anchors and delayed credit.}
A \emph{hard progress anchor} is created when a transition changes reward, inventory, a persistent object state, an unlocked region or exit, a persistent flag, or the terminal status. For example, adding an access code to the inventory yields an `obtain access code'' anchor, while changing a gate from locked to open yields an `open gate'' anchor. A \emph{discovery anchor} is created when the Actor first reaches a room, observes an object or exit, discovers a new admissible action, or encounters an available but untried opportunity. Anchor labels are instantiated from fixed templates rather than generated by an LLM.

To assign credit to preceding zero-reward actions, the Evolver defines

\begin{equation}
g_t=
\Delta r_t
+\lambda_I I_t
+\lambda_O O_t
+\lambda_E E_t
-\lambda_F F_t,
\label{eq:immediate_progress}
\end{equation}
where $I_t$, $O_t$, and $E_t$ indicate task-relevant inventory, object-state, and exploration progress, while $F_t$ indicates invalid actions, death, loops, or no progress. For a candidate segment ending at anchor step $t_a$, credit is propagated backward along the observed trajectory:
\begin{equation}
C_{t_a}=g_{t_a},
\qquad
C_t=g_t+\gamma C_{t+1},
\quad t<t_a,
\label{eq:delayed_credit}
\end{equation}

where $\gamma\in(0,1)$ is a step-level discount factor.

\paragraph{Experience-grounded prerequisite estimation.}
Delayed credit identifies temporally useful transitions but does not reveal the
conditions under which their actions are likely to succeed. The Evolver therefore
aggregates action attempts across episodes to identify facts that are empirically
associated with action success.

For each attempt $i$ of action $a$, let $y_i \in \{0,1\}$ denote whether the
action succeeds, and let $x_i^f \in \{0,1\}$ indicate whether fact $f$ has been
observed before the action is executed. We define the support of the two groups as
\begin{equation}
N_b(f,a)
=
\sum_{i \in \mathcal{D}_a}
\mathbb{I}[x_i^f=b],
\qquad b\in\{0,1\},
\label{eq:prerequisite_support}
\end{equation}
where $\mathcal{D}_a$ contains all observed attempts of action $a$ across
episodes. The empirical prerequisite score is then
\begin{equation}
\operatorname{Nec}(f,a)
=
\widehat{\Pr}(y=1\mid a,x^f=1)
-
\widehat{\Pr}(y=1\mid a,x^f=0).
\label{eq:necessity_score}
\end{equation}
A fact is included in the candidate prerequisite set only if
\begin{equation}
f \in \widehat{\operatorname{Pre}}(a)
\iff
\operatorname{Nec}(f,a)>\tau_{\mathrm{pre}}
\;\land\;
N_1(f,a)\geq n_{\min}
\;\land\;
N_0(f,a)\geq n_{\min}.
\label{eq:prerequisite_selection}
\end{equation}
Action--fact pairs that do not satisfy both support requirements, including
actions that have not been sufficiently attempted, remain unspecified rather
than being assigned a prerequisite relation. We consider only facts observed
before the action attempt, thereby excluding facts produced by the action itself.
Prerequisite candidates can additionally be supported by changes in admissible
actions, environment-provided failure reasons, and differences in action
availability across states.

Because this score is estimated from observational trajectories, it represents
an empirical association rather than a causal or logically necessary condition.
We therefore use it only to identify candidate prerequisites for subsequent
Strategy Map refinement.

\subsection{Fact-Grounded Dependency Linking and Walkthrough Extraction}
\label{sec:dependency_linking}

Prerequisite estimation determines which facts an action requires but does not identify which concrete transitions in the current trajectory produced those facts. Backward dependency slicing therefore starts from the anchor's completion facts and recursively searches the trajectory for the actions that produced them, yielding a compact, dependency-consistent action chain.

The positive effect set of transition $t$ is
\begin{equation}
\operatorname{Eff}_t=\operatorname{Diff}^{+}(s_t,s_{t+1}).
\label{eq:positive_effect}
\end{equation}
A transition with $\kappa_t=\kappa_{t+1}$ has $\operatorname{Eff}_t=\varnothing$ and is treated as a no-progress self-loop.

For an anchor, the unresolved fact set is initialized as
\begin{equation}
R\leftarrow\operatorname{CompleteFacts}(\mathrm{anchor}).
\label{eq:unresolved_initialization}
\end{equation}
The trajectory is scanned backward, and transition $t$ is retained when
\begin{equation}
\operatorname{Eff}_t\cap R\neq\varnothing.
\label{eq:slicing_condition}
\end{equation}
After retention, the unresolved set is updated by
\begin{equation}
R\leftarrow
\left(R\setminus\operatorname{Eff}_t\right)
\cup\operatorname{Pre}(a_t).
\label{eq:unresolved_update}
\end{equation}

When evidence is sparse, the Evolver conservatively removes only exact self-loops and explicit back-and-forth cycles, postponing further pruning until more episodes are available.

For instance, slicing backward from an open-gate anchor may retain \emph{unlock gate}, \emph{move to gate}, \emph{take code}, \emph{search desk}, and \emph{ask technician}, while removing unrelated observations and detours.

Each Walkthrough is represented compactly as
\begin{equation}
W_i=(g_i,D_i,e_i,\Pi_i,c_i,f_i,z_i,\eta_i),
\label{eq:walkthrough_definition}
\end{equation}
where $g_i$ is the local goal, $D_i$ is the prerequisite set, $e_i$ is the entry predicate, $\Pi_i$ is the ordered procedure, $c_i$ and $f_i$ are the completion and failure predicates, $z_i$ is the verification status, and $\eta_i$ stores scheduling statistics. The procedure is
\begin{equation}
\Pi_i=(p_{i,1},\ldots,p_{i,L_i}),
\qquad
p_{i,j}=(P_{i,j},a_{i,j},E_{i,j}),
\label{eq:walkthrough_steps}
\end{equation}

where $P_{i,j}$ is the step precondition, $a_{i,j}$ is the action, and $E_{i,j}$ is the expected effect. Thus, a step can be represented simply as ``technician visible $\rightarrow$ ask about gate $\rightarrow$ desk search enabled,'' rather than as a free-form textual memory.

The completion predicate is generated directly from the corresponding progress anchor, while the failure predicate captures terminal failure, irreversible loss of a required fact, or an unrecoverable prerequisite state. These explicit conditions make the induced procedure executable, resumable, and programmatically verifiable rather than merely descriptive.
 

\begin{table*}[t]
\caption{
(RQ1-A) \textbf{Overall performance comparison across three benchmarks.}
All methods are evaluated with the same backbone model within each block.
We highlight the \hlfirst{best} and \hlsecond{second-best} results.
$^{\dagger}$: No-learning Baseline.
$^{\star}$: Memory-based \& Reflection-based Methods.
$^{\diamond}$: Automated Prompt Optimization Methods.
}
\label{tab:main}
\centering
\scriptsize
\setlength{\tabcolsep}{0.5pt}

\resizebox{\textwidth}{!}{%
\begin{tabular}{@{}c|c|cc|cc|cc|cc@{}}
\Xhline{1.2pt}

\rowcolor{CadetBlue!20}
&
&
\multicolumn{2}{c|}{\textbf{WebShop}}
&
\multicolumn{2}{c|}{\textbf{J-TTL}}
&
\multicolumn{2}{c|}{\textbf{ScienceWorld}}
&
\multicolumn{2}{c}{\textbf{Avg.}}
\\

\rowcolor{CadetBlue!20}
\multirow{-2}{*}{\textbf{Backbone}}
&
\multirow{-2}{*}{\textbf{Method}}
&
\textbf{AUC} $\uparrow$
&
\textbf{Final-3} $\uparrow$
&
\textbf{AUC} $\uparrow$
&
\textbf{Final-3} $\uparrow$
&
\textbf{AUC} $\uparrow$
&
\textbf{Final-3} $\uparrow$
&
\textbf{AUC} $\uparrow$
&
\textbf{Final-3} $\uparrow$
\\

\Xhline{1.2pt}


\multirow{9}{*}{
    \rotatebox[origin=c]{90}{\modelname{Qwen3-32B}}
}

& \methodname{Static$^{\dagger}$}
& 0.4823\staticdelta
& 0.4706\staticdelta
& 0.0290\staticdelta
& 0.0191\staticdelta
& 0.1397\staticdelta
& 0.1150\staticdelta
& 0.2170\staticdelta
& 0.2016\staticdelta
\\

& \methodname{Memory$^{\star}$}
& \hlsecond{0.5403}\red{0.06}
& \hlsecond{0.5578}\red{0.09}
& 0.1327\red{0.10}
& 0.0583\red{0.04}
& -0.0273\blue{0.17}
& -0.0020\blue{0.12}
& 0.2152\blue{0.00}
& 0.2047\red{0.00}
\\

& \methodname{RAG$^{\star}$}
& 0.4510\blue{0.03}
& 0.4321\blue{0.04}
& 0.0263\blue{0.00}
& 0.0180\blue{0.00}
& 0.1200\blue{0.02}
& 0.1170\red{0.00}
& 0.1991\blue{0.02}
& 0.1890\blue{0.01}
\\

& \methodname{Summary$^{\star}$}
& 0.4680\blue{0.01}
& 0.4943\red{0.02}
& 0.1461\red{0.12}
& 0.0654\red{0.05}
& 0.0759\blue{0.06}
& 0.0624\blue{0.05}
& 0.2300\red{0.01}
& 0.2074\red{0.01}
\\

& \methodname{Reflexion$^{\star}$}
& 0.4859\red{0.00}
& 0.4728\red{0.00}
& \hlsecond{0.2380}\red{0.21}
& \hlsecond{0.1870}\red{0.17}
& 0.1388\blue{0.00}
& 0.1513\red{0.04}
& \hlsecond{0.2876}\red{0.07}
& \hlsecond{0.2704}\red{0.07}
\\

& \methodname{EvoTest$^{\diamond}$}
& 0.4889\red{0.01}
& 0.5094\red{0.04}
& 0.1562\red{0.13}
& 0.0718\red{0.05}
& 0.1507\red{0.01}
& 0.1446\red{0.03}
& 0.2653\red{0.05}
& 0.2419\red{0.04}
\\

& \methodname{ACE$^{\diamond}$}
& 0.2128\blue{0.27}
& 0.2317\blue{0.24}
& 0.1855\red{0.16}
& 0.0768\red{0.06}
& \hlsecond{0.1530}\red{0.01}
& \hlsecond{0.1602}\red{0.05}
& 0.1838\blue{0.03}
& 0.1563\blue{0.05}
\\

& \methodname{APEX$^{\diamond}$}
& 0.5034\red{0.02}
& 0.5412\red{0.07}
& 0.1668\red{0.14}
& \hlsecond{0.1870}\red{0.17}
& 0.0336\blue{0.11}
& 0.0028\blue{0.11}
& 0.2346\red{0.02}
& 0.2437\red{0.04}
\\

& \method
& \hlfirst{0.5421}\red{0.06}
& \hlfirst{0.6047}\red{0.13}
& \hlfirst{0.2410}\red{0.21}
& \hlfirst{0.2833}\red{0.26}
& \hlfirst{0.1884}\red{0.05}
& \hlfirst{0.2186}\red{0.10}
& \hlfirst{0.3238}\red{0.11}
& \hlfirst{0.3689}\red{0.17}
\\

\midrule


\multirow{9}{*}{
    \rotatebox[origin=c]{90}{\modelname{Mistral-Small-3.2-24B}}
}

& \methodname{Static$^{\dagger}$}
& 0.3974\staticdelta
& 0.3867\staticdelta
& 0.0398\staticdelta
& 0.0309\staticdelta
& 0.1626\staticdelta
& 0.1333\staticdelta
& 0.1999\staticdelta
& 0.1836\staticdelta
\\

& \methodname{Memory$^{\star}$}
& 0.4512\red{0.05}
& 0.5549\red{0.17}
& 0.1212\red{0.08}
& 0.0847\red{0.05}
& 0.1479\blue{0.01}
& 0.1378\red{0.00}
& 0.2401\red{0.04}
& 0.2591\red{0.08}
\\

& \methodname{RAG$^{\star}$}
& 0.3386\blue{0.06}
& 0.3524\blue{0.03}
& 0.0544\red{0.01}
& 0.0740\red{0.04}
& 0.1215\blue{0.04}
& 0.1511\red{0.02}
& 0.1715\blue{0.03}
& 0.1925\red{0.01}
\\

& \methodname{Summary$^{\star}$}
& 0.4149\red{0.02}
& 0.5118\red{0.13}
& 0.1610\red{0.12}
& 0.1053\red{0.07}
& 0.1658\red{0.00}
& 0.1767\red{0.04}
& 0.2472\red{0.05}
& 0.2646\red{0.08}
\\

& \methodname{Reflexion$^{\star}$}
& 0.4461\red{0.05}
& 0.4295\red{0.04}
& 0.2078\red{0.17}
& 0.1624\red{0.13}
& 0.1628\red{0.00}
& \hlsecond{0.2396}\red{0.11}
& 0.2722\red{0.07}
& 0.2772\red{0.09}
\\

& \methodname{EvoTest$^{\diamond}$}
& \hlsecond{0.4627}\red{0.07}
& 0.5268\red{0.14}
& 0.2073\red{0.17}
& 0.1608\red{0.13}
& 0.1630\red{0.00}
& 0.1914\red{0.06}
& 0.2777\red{0.08}
& 0.2930\red{0.11}
\\

& \methodname{ACE$^{\diamond}$}
& 0.2875\blue{0.11}
& 0.2913\blue{0.10}
& 0.1847\red{0.14}
& 0.1470\red{0.12}
& 0.1889\red{0.03}
& 0.2188\red{0.09}
& 0.2204\red{0.02}
& 0.2190\red{0.04}
\\

& \methodname{APEX$^{\diamond}$}
& 0.4318\red{0.03}
& \hlsecond{0.5682}\red{0.18}
& \hlsecond{0.2250}\red{0.19}
& \hlsecond{0.1808}\red{0.15}
& \hlsecond{0.1927}\red{0.03}
& 0.2055\red{0.07}
& \hlsecond{0.2832}\red{0.08}
& \hlsecond{0.3182}\red{0.13}
\\

& \method
& \hlfirst{0.4896}\red{0.09}
& \hlfirst{0.5726}\red{0.19}
& \hlfirst{0.3742}\red{0.33}
& \hlfirst{0.3195}\red{0.29}
& \hlfirst{0.2243}\red{0.06}
& \hlfirst{0.2476}\red{0.11}
& \hlfirst{0.3627}\red{0.16}
& \hlfirst{0.3799}\red{0.20}
\\

\midrule


\multirow{9}{*}{
    \rotatebox[origin=c]{90}{\modelname{GPT-OSS-20B}}
}

& \methodname{Static$^{\dagger}$}
& 0.4523\staticdelta
& 0.4189\staticdelta
& 0.0253\staticdelta
& 0.0130\staticdelta
& 0.1731\staticdelta
& \hlsecond{0.1949}\staticdelta
& 0.2169\staticdelta
& 0.2089\staticdelta
\\

& \methodname{Memory$^{\star}$}
& \hlsecond{0.5123}\red{0.06}
& 0.5567\red{0.14}
& 0.0567\red{0.03}
& 0.0548\red{0.04}
& 0.1530\blue{0.02}
& 0.1661\blue{0.03}
& 0.2407\red{0.02}
& 0.2592\red{0.05}
\\

& \methodname{RAG$^{\star}$}
& 0.4210\blue{0.03}
& 0.3812\blue{0.04}
& 0.0076\blue{0.02}
& 0.0220\red{0.01}
& 0.1058\blue{0.07}
& 0.0996\blue{0.10}
& 0.1781\blue{0.04}
& 0.1676\blue{0.04}
\\

& \methodname{Summary$^{\star}$}
& 0.4380\blue{0.01}
& 0.4915\red{0.07}
& 0.1155\red{0.09}
& 0.0622\red{0.05}
& 0.1497\blue{0.02}
& 0.1486\blue{0.05}
& 0.2344\red{0.02}
& 0.2341\red{0.03}
\\

& \methodname{Reflexion$^{\star}$}
& 0.4529\red{0.00}
& 0.4263\red{0.01}
& 0.1105\red{0.09}
& 0.0687\red{0.06}
& 0.1732\red{0.00}
& 0.1877\blue{0.01}
& 0.2455\red{0.03}
& 0.2276\red{0.02}
\\

& \methodname{EvoTest$^{\diamond}$}
& 0.4529\red{0.00}
& \hlsecond{0.5778}\red{0.16}
& \hlsecond{0.1375}\red{0.11}
& \hlsecond{0.0971}\red{0.08}
& 0.1287\blue{0.04}
& 0.1544\blue{0.04}
& 0.2397\red{0.02}
& \hlsecond{0.2764}\red{0.07}
\\

& \methodname{ACE$^{\diamond}$}
& 0.1728\blue{0.28}
& 0.1632\blue{0.26}
& 0.0913\red{0.07}
& 0.0413\red{0.03}
& 0.1401\blue{0.03}
& 0.1394\blue{0.06}
& 0.1347\blue{0.08}
& 0.1146\blue{0.09}
\\

& \methodname{APEX$^{\diamond}$}
& 0.4654\red{0.01}
& 0.5441\red{0.13}
& 0.0906\red{0.07}
& 0.0613\red{0.05}
& \hlsecond{0.1972}\red{0.02}
& -0.0833\blue{0.28}
& 0.2511\red{0.03}
& 0.1740\blue{0.03}
\\

& \method
& \hlfirst{0.5143}\red{0.06}
& \hlfirst{0.5960}\red{0.18}
& \hlfirst{0.3783}\red{0.35}
& \hlfirst{0.4927}\red{0.48}
& \hlfirst{0.2527}\red{0.08}
& \hlfirst{0.3113}\red{0.12}
& \hlfirst{0.3818}\red{0.16}
& \hlfirst{0.4667}\red{0.26}
\\

\bottomrule
\end{tabular}%
}
\end{table*}

\section{Experiments} 
\label{sec:experiments}

\begin{figure*}[t]
    \centering
    \includegraphics[width=\textwidth]{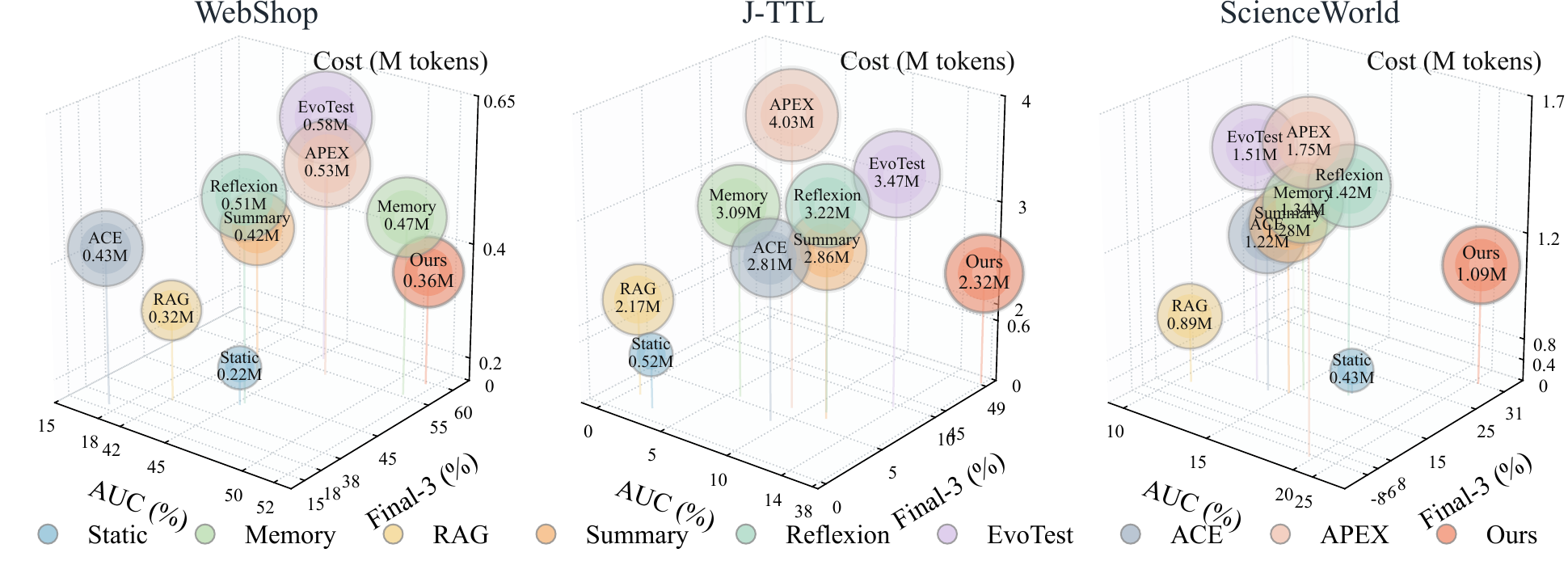}
    \caption{
        (RQ1-B) \textbf{Performance--cost comparison using \modelname{GPT-OSS-20B}.}
        All methods use the same backbone and are evaluated under the same
        ten-episode interaction budget.
        Total Tokens include both input and output tokens consumed over all episodes
        and are reported in millions (M).
        We highlight the \hlfirst{best} and \hlsecond{second-best} performance.
    }
    \label{fig:performance_cost}
\end{figure*}


In this section, we conduct extensive experiments to answer the following research questions: \textbf{(RQ1)}: How does \method compare with strong test-time learning and memory baselines? \textbf{(RQ2)}: Does each construction mechanism correctly perform its intended function? \textbf{(RQ3)} How do Walkthrough representation and dependency-aware scheduling affect cross-episode experience reuse?  \textbf{(RQ4)}: How robust is \method, and under what conditions does its advantage diminish?
%

\subsection{Experimental Setup}

\paragraph{Datasets and Benchmarks. }To thoroughly evaluate the effectiveness of \method, we adopt three widely-adopted benchmarks across three domains: \textbf{(1) Interactive Fiction}, including  J-TTL~\citep{hausknecht2020interactive, he2026evotest}. \textbf{(2) E-commerce}, including WebShop~\citep{yao2022webshop}. \textbf{(3) Scientific Reasoning}, including ScienceWorld~\citep{wang2022scienceworld}. Further experimental details appear in the Appendix.

\paragraph{Baselines.}
We evaluate eight baselines:
(1) \modelname{Static};
(2) \modelname{Memory}~\citep{he2026evotest};
(3) \modelname{RAG}~\citep{he2026evotest};
(4) \modelname{Summary}~\citep{he2026evotest};
(5) \modelname{Reflexion}~\citep{shinn2023reflexion};
(6) \modelname{EvoTest}~\citep{he2026evotest};
(7) \modelname{ACE}~\citep{zhang2025agentic}; and
(8) \modelname{APEX}~\citep{li2026apex}.
\paragraph{LLM Backbones. }For instantiating these frameworks,
we adopt three open-source LLMs, \modelname{Qwen3-32B}, \modelname{Mistral-Small-3.2-24B}, and \modelname{GPT-OSS-20B}. 

\paragraph{Metrics.}
We run each game for $E=10$ episodes and report:
(1) \textbf{Final-$3$}, the average normalized score over the last three episodes;
(2) \textbf{AUC}, following \citet{he2026evotest}, defined as
$\mathrm{AUC}=\frac{1}{E}\sum_{e=1}^{E}\frac{R(e)}{R_{\max}}$,
where $R(e)$ is the score obtained in episode $e$ and $R_{\max}$ is the maximum score achievable in a single episode;
(3) \textbf{Replay Success} (\textbf{RS}), the fraction of attempted Walkthrough executions that reach their specified completion conditions, defined as
$\mathrm{RS}=\frac{1}{N_{\mathrm{rep}}}\sum_{i=1}^{N_{\mathrm{rep}}}\mathbb{I}[c_i=1]$,
where $N_{\mathrm{rep}}$ is the total number of Walkthrough replay attempts and $c_i=1$ if the $i$-th replay reaches its completion condition before triggering a failure condition or exhausting its execution budget, and $c_i=0$ otherwise; and
(4) \textbf{Token Cost}, the total number of input and output tokens consumed across all episodes, defined as
$\mathrm{Cost}=\sum_{e=1}^{E}\left(T^{\mathrm{in}}_e+T^{\mathrm{out}}_e\right)$.

\paragraph{Parameter Configurations.}
For all experiments, we set the LLM temperature to $0.8$. Each task is run for $10$ episodes, with each episode limited to at most $150$ environment steps.


\subsection{Main Results (RQ1)}


\paragraph{Takeaway~$1$: \method achieves the best performance across tasks and backbone models.}
As shown in Table~\ref{tab:main}, compared with the strongest baseline in the Avg. column for each backbone and metric, \method improves the average AUC and Final-3 by $30.0\%$ and $40.5\%$, respectively. On \modelname{Qwen3-32B}, \modelname{Mistral-Small-3.2-24B}, and \modelname{GPT-OSS-20B}, the AUC gains are $12.6\%$, $28.1\%$, and $52.1\%$, while the Final-3 gains are $36.4\%$, $19.4\%$, and $68.8\%$, respectively. The improvements are most pronounced on J-TTL, where \method increases AUC and Final-$3$ by $65.4\%$ and $135.6\%$ on average.

\paragraph{Takeaway~$2$: \method{} achieves its performance gains without incurring excessive inference cost.}
As shown in Figure~\ref{fig:performance_cost}, \method{} consumes only $0.3564$M, $2.3231$M, and $1.0896$M tokens on WebShop, J-TTL, and ScienceWorld, respectively. Its token consumption is consistently lower than that of Memory, Summary, Reflexion, EvoTest, ACE, and APEX across all three benchmarks. More importantly, \method{} is also more efficient than the AUC-best baseline in each setting: it uses $23.6\%$ fewer tokens than Memory on WebShop, $33.1\%$ fewer than EvoTest on J-TTL, and $37.8\%$ fewer than APEX on ScienceWorld. These results show that the improvements of \method{} arise from more effective cross-episode experience reuse rather than increased inference expenditure.

\subsection{Ablation Study (RQ2)}

\begin{figure}[t]
\centering

\begin{minipage}[t]{0.58\textwidth}
\vspace{0pt}
\centering

\captionof{table}{
(RQ2) \textbf{Nested ablation of the three Walkthrough construction stages illustrated in Figure~\ref{fig:overview}, evaluated on ScienceWorld and J-TTL using \modelname{GPT-OSS-20B}.}
}
\label{tab:rq2_ablation}

\scriptsize
\setlength{\tabcolsep}{0.5pt}
\renewcommand{\arraystretch}{1}

\resizebox{\linewidth}{!}{%
\begin{tabular}{@{}ccc|cc|cc@{}}
\Xhline{1.2pt}

\rowcolor{CadetBlue!20}
&
&
&
\multicolumn{2}{c|}{\textbf{ScienceWorld}}
&
\multicolumn{2}{c}{\textbf{J-TTL}}
\\

\rowcolor{CadetBlue!20}
\multirow{-2}{*}{
    \shortstack[c]{\textbf{Del.}\\\textbf{Cred.}}
}
&
\multirow{-2}{*}{
    \shortstack[c]{\textbf{Prereq.}\\\textbf{Est.}}
}
&
\multirow{-2}{*}{
    \shortstack[c]{\textbf{Back.}\\\textbf{Slic.}}
}
&
\textbf{AUC} $\uparrow$
&
\textbf{Final-$3$} $\uparrow$
&
\textbf{AUC} $\uparrow$
&
\textbf{Final-$3$} $\uparrow$
\\

\Xhline{1.2pt}

$\checkmark$
& $\checkmark$
& $\checkmark$
& 0.2527 \red{0.00}
& 0.3113 \red{0.00}
& 0.3783 \red{0.00}
& 0.4927 \red{0.00}
\\

\rowcolor{gray!10}
$\checkmark$
& $\checkmark$
& $\times$
& 0.1841 \blue{0.07}
& 0.2284 \blue{0.08}
& 0.3116 \blue{0.07}
& 0.4072 \blue{0.09}
\\

$\checkmark$
& $\times$
& $\times$
& 0.1453 \blue{0.04}
& 0.1815 \blue{0.05}
& 0.2398 \blue{0.07}
& 0.3158 \blue{0.09}
\\

\rowcolor{gray!10}
$\times$
& $\times$
& $\times$
& 0.1072 \blue{0.04}
& 0.1346 \blue{0.05}
& 0.1664 \blue{0.07}
& 0.2183 \blue{0.10}
\\

\Xhline{1.2pt}
\end{tabular}%
}

\end{minipage}
\hfill
\begin{minipage}[t]{0.39\textwidth}
\vspace{0pt}
\centering

\includegraphics[
    width=\linewidth
]{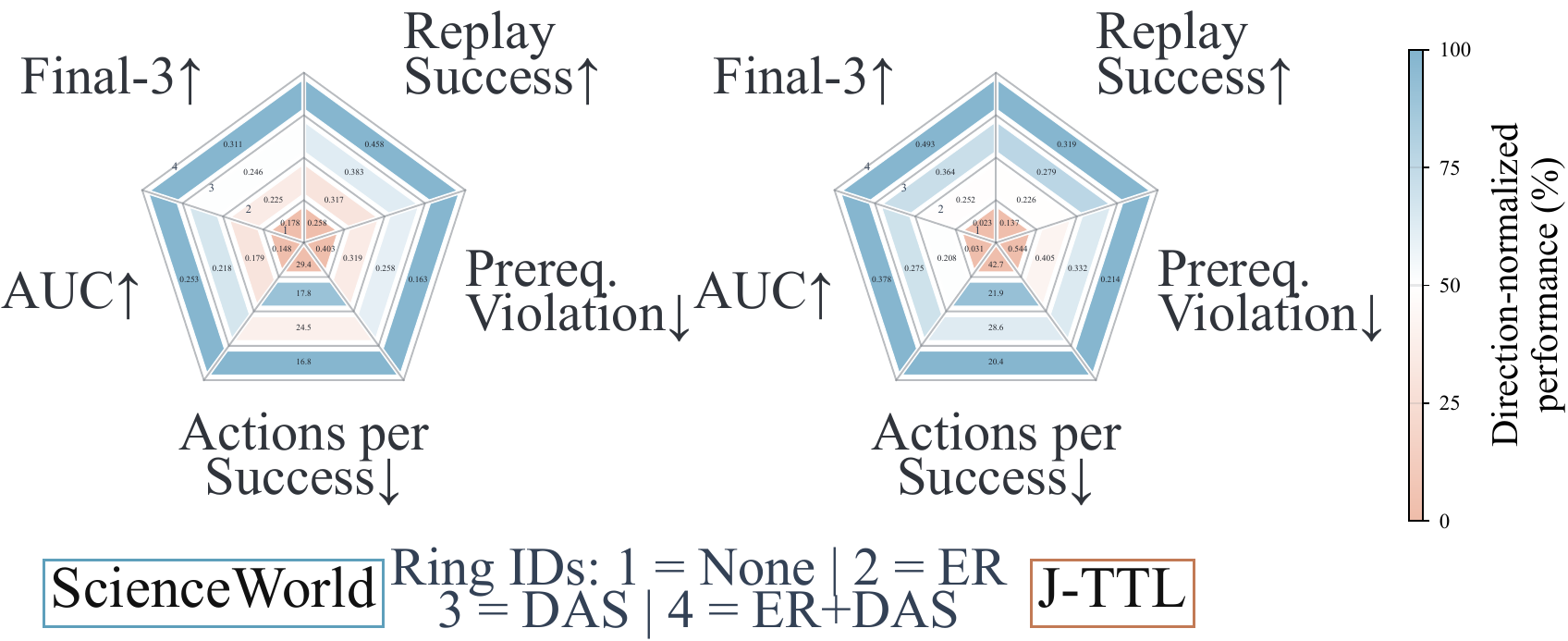}

\captionof{figure}{
(RQ3) \textbf{Effects of executable Walkthrough representation and
dependency-aware scheduling on cross-episode experience reuse.}
All variants use the same nodes, action sequences,
prerequisite annotations, dependency edges, memory snapshots,
and interaction budgets.
}
\label{fig:rq3_execution_scheduling}

\end{minipage}

\end{figure}

\paragraph{Takeaway~$3$: All three construction stages provide complementary and cumulative gains in end-to-end performance.}
As shown in Table~\ref{tab:rq2_ablation}, adding Delayed Credit to Immediate Progress Only improves AUC by $0.0381$ and Final-$3$ by $0.0469$ on ScienceWorld, while improving AUC by $0.0734$ and Final-$3$ by $0.0975$ on J-TTL. Prerequisite Estimation provides AUC and Final-$3$ gains of $0.0388$ and $0.0469$ on ScienceWorld, and $0.0718$ and $0.0914$ on J-TTL. Backward Dependency Slicing further increases AUC by $0.0686$ and Final-$3$ by $0.0829$ on ScienceWorld, and increases AUC by $0.0667$ and Final-$3$ by $0.0855$ on J-TTL. Overall, the full method exceeds Immediate Progress Only by $0.1455$ in AUC and $0.1767$ in Final-$3$ on ScienceWorld, and by $0.2119$ in AUC and $0.2744$ in Final-$3$ on J-TTL. These cumulative gains show that candidate recovery, prerequisite induction, and dependency-based pruning jointly improve performance.

\subsection{Case Study(RQ3)}

\paragraph{Takeaway~$4$: Executable representation and dependency-aware scheduling improve execution efficiency and dependency consistency, respectively, and provide complementary benefits.}
As shown in Figure~\ref{fig:rq3_execution_scheduling}, enabling either mechanism alone improves every metric on both ScienceWorld and J-TTL. Executable representation primarily reduces execution cost, lowering Actions per Success from $29.4$ to $17.8$ on ScienceWorld and from $42.7$ to $21.9$ on J-TTL. Dependency-aware scheduling more strongly improves reuse validity: it increases Replay Success by $0.125$ and $0.142$, while reducing Prerequisite Violation by $0.145$ and $0.212$, on ScienceWorld and J-TTL, respectively. Each mechanism provides further gains when added on top of the other, and the full method achieves the best results across all metrics. 

\begin{wrapfigure}{r}{0.48\textwidth}
    \vspace{-3mm}
    \centering
    \includegraphics[
        width=\linewidth
    ]{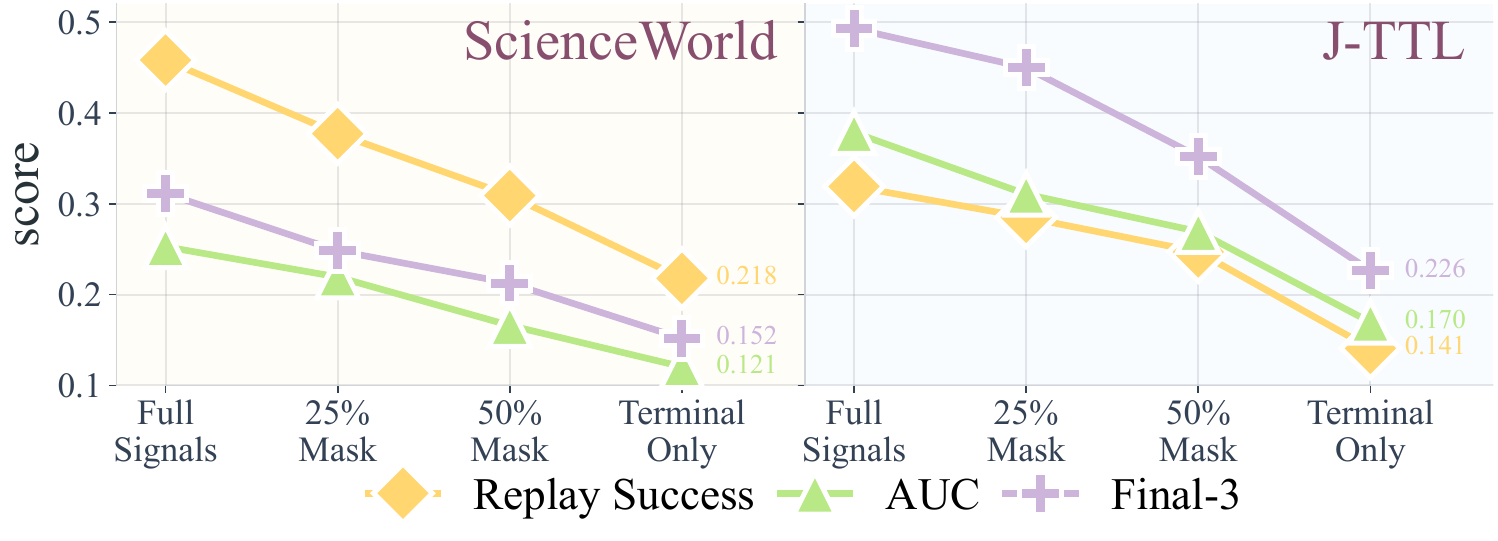}
    \caption{
        (RQ4-A) \textbf{Robustness to progress-signal masking on ScienceWorld and J-TTL using \modelname{GPT-OSS-20B}.}
    }
    \label{fig:rq4a_progress_signal_masking}
    \vspace{-4mm}
\end{wrapfigure}

\subsection{Robustness and Sensitivity Study (RQ4)}
\paragraph{Takeaway~$5$: Delayed Credit remains effective under partial masking of non-terminal progress signals, but relying only on terminal rewards substantially degrades Walkthrough executability and downstream performance.}
As shown in Figure~\ref{fig:rq4a_progress_signal_masking}, masking $25\%$ of the non-terminal progress signals reduces Replay Success by only $0.054$--$0.061$, while AUC and Final-$3$ decrease by $0.0333$--$0.0578$ and $0.0430$--$0.0731$, respectively. All three metrics decline consistently as more intermediate signals are removed. When only terminal rewards are retained, Replay Success decreases by $0.178$--$0.240$, accompanied by substantially larger reductions in AUC and Final-$3$. These results show that Delayed Credit tolerates partially missing progress evidence, whereas intermediate state changes remain essential for constructing executable Walkthroughs and improving subsequent episodes.

\begin{wrapfigure}{r}{0.48\textwidth}
    \vspace{-3mm}
    \centering
    \includegraphics[
        width=\linewidth
    ]{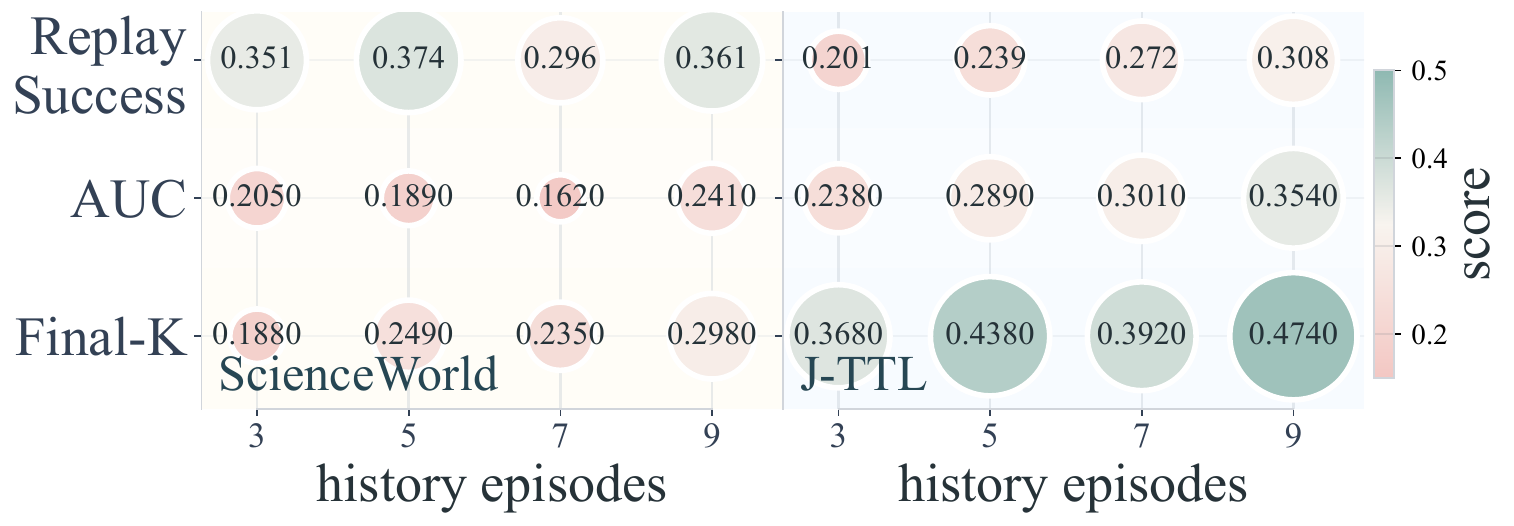}
    \caption{
        (RQ4-B) \textbf{Cross-episode evidence sample efficiency on ScienceWorld and J-TTL using \modelname{GPT-OSS-20B}.}
    }
    \label{fig:rq4c_sample_efficiency}
    \vspace{-4mm}
\end{wrapfigure}

\paragraph{Takeaway~$6$: Additional cross-episode evidence improves endpoint performance within the ten-episode interaction budget.}
As shown in Figure~\ref{fig:rq4c_sample_efficiency}, increasing the available history from $3$ to $9$ episodes improves Replay Success, AUC, and Final-3 by $0.010$, $0.036$, and $0.110$ on ScienceWorld, and by $0.107$, $0.116$, and $0.106$ on J-TTL, respectively. The intermediate results are not strictly monotonic, but the nine-episode setting outperforms the three-episode setting on every metric and both benchmarks. These results show that Prerequisite Estimation can benefit from additional success and failure observations within the practical ten-episode interaction budget.

\begin{wrapfigure}{r}{0.48\textwidth}
    \vspace{-3mm}
    \centering
    \includegraphics[
        width=\linewidth
    ]{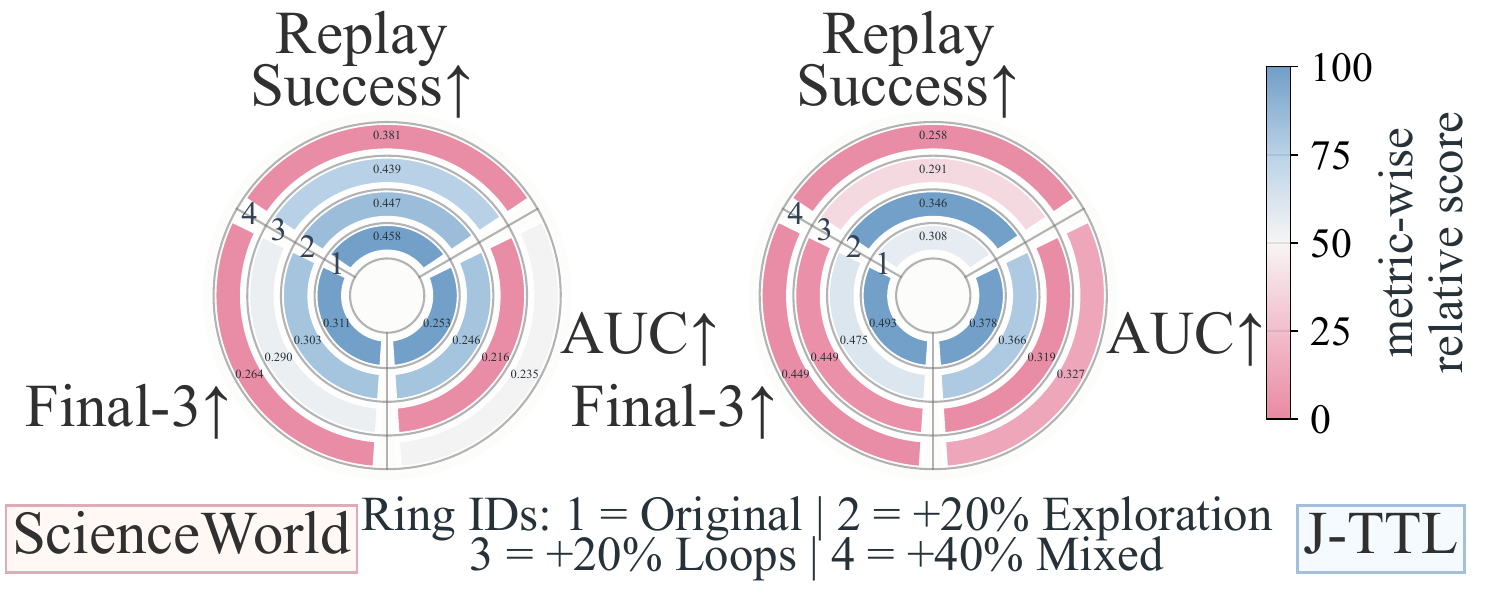}
    \caption{
        (RQ4-C) \textbf{Robustness to trajectory noise on ScienceWorld and J-TTL using \modelname{GPT-OSS-20B}.}
    }
    \label{fig:rq4b_trajectory_noise}
    \vspace{-4mm}
\end{wrapfigure}

\paragraph{Takeaway~$7$: \method is robust to trajectory noise, retaining most of its clean-trajectory performance even under substantial structured redundancy.}
As shown in Figure~\ref{fig:rq4b_trajectory_noise}, with $20\%$ isolated exploratory actions, the metrics that decrease retain $96.3\%$--$97.6\%$ of their original performance across both benchmarks, while Replay Success on J-TTL increases from $0.308$ to $0.346$. Under the more challenging setting with $20\%$ loops and backtracking, the method preserves $84.4\%$--$95.9\%$ of its clean-trajectory performance. Even when mixed redundancy reaches $40\%$, it retains $83.2\%$--$92.9\%$ of the corresponding clean values. These results demonstrate that the Walkthrough construction pipeline effectively suppresses interference from exploration, loops, and backtracking while maintaining stable executability and downstream gains.

\section{Conclusion}


We introduce \method to learn executable Walkthroughs from noisy, sparse-reward trajectories for reliable experience reuse. \method combines delayed-credit propagation, fact-grounded prerequisite estimation, and backward dependency slicing to extract progress-enabling action chains. Empirically, it outperforms the strongest baselines by $30.0\%$ in AUC and $40.5\%$ in Final-$3$ with lower token cost. Ablations and robustness studies confirm complementary gains from all stages and resilience to trajectory noise. These results establish state-conditioned executable procedures as a practical memory unit for efficient, verifiable, and resumable long-horizon agents.



\bibliography{iclr2027_conference}
\bibliographystyle{iclr2027_conference}

\end{document}